# Ontology, Ontologies, and Science

## Gary H. Merrill

ABSTRACT   Philosophers frequently struggle with the relation of metaphysics to the everyday world, with its practical value, and with its relation to empirical science.  This paper distinguishes several different models of the relation between philosophical ontology and applied (scientific) ontology that have been advanced in the history of philosophy.  Adoption of a *strong participation model* for the philosophical ontologist in science is urged, and requirements and consequences of the participation model are explored.  This approach provides both a principled view and justification of the role of the philosophical ontologist in contemporary empirical science as well as guidelines for integrating philosophers and philosophical contributions into the practice of science.

# Introduction

Metaphysicians, when explaining or justifying their calling, tend to be a mournful and defensive lot while at the same time extolling the intellectual, moral, and spiritual virtues of metaphysics and its practice.  A classic example is found in Russell's *The Problems of Philosophy* where he argues that philosophy as a discipline is not quite as fruitless as it may appear:

> Philosophy, like all other studies, aims primarily at knowledge.... But it cannot be maintained that philosophy has had any very great measure of success in its attempts to provide definite answers to its questions....  It is true that this is partly accounted for by the fact that as soon as definite knowledge concerning any subject becomes possible, this subject ceases to be called philosophy, and becomes a separate science....
>
> Philosophy is to be studied, not for the sake of any definite answers to its questions since no definite answers can, as a rule, be known to be true, but rather for the sake of the questions themselves; because these questions enlarge our conception of what is possible, enrich our intellectual imagination and diminish the dogmatic assurance which closes the mind against speculation; but above all because, through the greatness of the universe which philosophy contemplates, the mind also is rendered great, and becomes capable of that union with the universe which constitutes its highest good. (Russell, 2010, p. 98)[1]

This shadows, in somewhat more flowery prose, Hume's sentiments in the *Enquiry* when he writes that

---

[1]     While Russell speaks here generally of philosophy rather than more narrowly of metaphysics, and although the focus in *The Problems of Philosophy* is intended to be primarily on epistemology, the context of his remarks at this point makes it clear that he intends them to encompass metaphysics and (even more specifically) metaphysical and ontological questions raised and settled in the empirical sciences.

> And though a philosopher may live remote from business, the genius of
> philosophy, if carefully cultivated by several, must gradually diffuse itself
> throughout the whole society, and bestow a similar correctness on every art and
> calling. (Hume 2007, p. 7)

A related theme, that metaphysics is, or involves, a specific "way of thinking", is expressed to some degree by both Hume and Russell, and by Richard Taylor in the opening chapters of his *Metaphysics* where he (like Russell) urges the view that metaphysics yields not knowledge, but *understanding* and *wisdom*. Taylor sees the practice of metaphysics as centered on *problems* — problems that arise from our interactions with, and comtemplations of, the world as we attempt to understand the "data" (his word) that we encounter within it. He echoes Russell in saying of metaphysics that

> Metaphysics, in fact, promises no *knowledge* of anything. If knowledge itself is
> what you seek, be grateful for empirical science, for you will never find it in
> metaphysics. (Taylor 1992, pp. 6-7)

These views are of metaphysics writ broadly and large — while at the same time exhibiting an undercurrent of defensiveness and insecurity. The situation for admitted metaphysicians is hardly improved by attacks from fellow philosophers, including Hume himself ("cast it into the flames") and the early Carnap ("alleged statements in this domain are entirely meaningless"), and perhaps the fear of sterility or meaninglessness is what historically has driven metaphysicians to draw such a stark line of demarcation between themselves and scientists: if you are not striving to attain knowledge, then you can hardly be faulted for your failure in attaining it.

Still, some philosophers feel that the connection between philosophical ontology and the ontologies of science is more immediate and direct. In his *New Essays on Human Understanding*, Leibniz remarks that

> The art of ranking things in genera and species is of no small importance and
> very much assists our judgment as well as our memory. You know how much it
> matters in botany, not to mention animals and other substances, or again 'moral'
> and 'notional' entities as some call them. Order largely depends on it, and many
> good authors write in such a way that their whole account could be divided and
> subdivided according to a procedure related to genera and species. This helps
> one not merely to retain things, but also to find them. And those who have laid
> out all sorts of notions under certain headings or categories have done something
> very useful. (Leibniz 1996, pp. 179-180)

The very useful thing to which Leibniz refers is of course one of the primary contributions that the philosophical ontologist can make to the sciences. This consists in the creation of a set of "abstract ideas" (let us think of them as ontological categories) together with a set of names by which they may be referenced, and a system in which these categories are arrayed by means of relations (some of them hierarchical). Moreover, in the context of discussing the value of such categorization in geometry, Leibniz remarks with some prescience that

> To these two kinds of arrangement [synthetic and analytic] we must add a third.
> It is classification by terms, and really all it produces is a kind of Inventory.
> The latter could be systematic, with the terms being ordered according to certain
> categories shared by all peoples, or it could have an alphabetical order within the

> accepted language of the learned world. ... And there is even more reason why these inventories should be more useful in the other sciences, where the art of reasoning has less power, and they are utterly necessary in medicine above all. (Leibniz 1996, p. 382)

Thus has Leibniz, from a distance of more than 300 years, characterized much of the work being done today in the domain of informatics, and especially medical informatics — mentioning explicitly the value of systematic classification in domains that we would now think of as knowledge representation, knowledge management, information retrieval, and inferencing — and Leibniz sees it as the work of the philosopher.

In a similar vein, Peirce comments that

> The task of classifying all the words of language, or what's the same thing, all the ideas that seek expression, is the most stupendous of logical tasks. Anybody but the most accomplished logician must break down in it utterly; and even for the strongest man, it is the severest possible tax on the logical equipment and faculty. (Peirce 1897)

The view here is consistent and uniform: that the philosopher's job is the construction of a system (or of systems) of categories to be used in describing and understanding the world around us. And Leibniz explicitly recognizes the value of ontology in the sciences, holding that it is in fact "utterly necessary" in the science of medicine.

More recently Thomas Hofweber has taken up a theme similar to Russell's and Taylor's in his "Ambitious, yet modest, metaphysics" (Hofweber 2009). He is concerned about "how to defend ontology as a philosophical discipline," particularly against what is fundamentally the Russellian view that the questions that metaphysics attempts to answer are, in the long run, better answered by the sciences and that therefore (Hofweber) "There is nothing left to do for philosophy ...". To counter this lament, Hofweber turns his attention to ontology (as a particularly well-delimited sub-domain of metaphysics) and advances a proposal concerning what the role of the ontologist should be, and what the value of that role is.

Hofweber believes that we may "save metaphysics" and more particularly "our beloved discipline of ontology"; and that the way to do this is to convert the ontologist to a linguistic analyst who will settle existence claims in "overlap cases" (where both the sciences and philosophy "have an interest in the same subject matter"). The settlement of such a claim as "Numbers exist." will then be accomplished by the proper examination of number talk in natural language; and this will yield a determination as to whether number terms are used in an internal (non-referential) sense — in which case numbers will be seen not to exist — or an external (referential) sense — in which case numbers will be seen as existing. I think there is much wrong with this view and approach of Hofweber's, but for our immediate concerns it is sufficient to note that it is an essentially defensive position that appears more timid than ambitious and seeks to reduce the risk that the philosopher will encroach on the territory of the scientist. In doing this, the Hofweberian philosopher is to retreat from ontology to meta-ontology and concede that philosophical ontology lacks relevance to other disciplines; and this approach ensures that ontology as practiced by the philosopher will be of interest only to other philosophers.[2]

A different course is taken by Dale Jacquette in which he carefully distinguishes between *pure philosophical ontology* and *applied scientific ontology*. For Jacquette, pure philosophical ontology is characterized by the fundamental question of what it *means* for something to exist — of "the precise meaning of the words 'being', 'to be', 'exist', 'existence`, to be 'real', 'actual', 'present', 'manifest', and like cognates."[3] Answering this question, for Jacquette, necessarily precedes investigating any questions of applied scientific ontology and we must have this answer before we can proceed in any intelligible way with the questions raised by scientific ontology. Applied scientific ontology, on the other hand, is concerned with building a "conceptual model" of what it means for something to exist, recommending a preferred existence domain (numbers, sets, atoms, fields, genes, ...), and *applying the definition of being* that emerges from pure philosophical ontology. Jacquette urges that only by achieving such a definition can coherent analyses and comparisons of scientific ontologies be made, and that such a definition is a necessary condition for rendering the work of the scientific ontologist intelligible. But although Jacquette believes that "The study of ontology belongs squarely to philosophy" and sees successful ontology as being an "integration" of philosophical ontology and scientific ontology where the latter is responsible for "grafting an appropriate preferred existence domain onto a satisfactory analysis of the concept of being", it is not at all clear what the *instrumental* role of philosophical ontology in this task would be: of precisely *what* the philosophical ontologist would affect in science and *how* this would be accomplished (cf. Jacquette 2002, pp. 275-280).

---

[2] This is most evident in Hofweber's discussion of the domain of ontology and the existence of numbers (Hofweber 2009, pp. 283-284). Here, it becomes clear that in the case of what Hofweber sees as the interesting philosophical questions "left open by science", these are left open because science simply does not care about their answers — or at least the answers that a philosopher would provide. No matter what answer the philosopher may offer to the question "Do numbers exist?", this will be of no consequence to the scientist. David Manley makes a related point in his "Introduction" to Chalmers 2009 when he says "On Hofweber's view, the claims of metaphysics do not conflict with the claims forthcoming from such other disciplines." (Chalmers, et al, p. 35) But if they do not conflict in such cases, they cannot be relevant to (or certainly cannot contribute to) those other disciplines and so will be of no interest to them — a somewhat odd incongruity if the original point was to deal with cases which overlapped in the sense that both scientists and philosophers have an interest in them, and to deal with them in part *because* of such an overlap of interest. This exposes a difficulty with Hofweber's notion of *overlap* and the weight it must support in his proposal. It can support that weight only if the overlap involves a *shared interest* or *shared sense*. The only sense of "shared interest" here is that both the scientist and the philosopher would agree that a particular statement (e.g., "Protons exist") is of interest. But the degree of sharing appears to be purely syntactic (agreement that the truth of this sentence is of interest), and not semantic (agreement on the semantics or truth conditions whose satisfaction would render the sentence true) or pragmatic (what are the consequences of the truth or falsity of the sentence). Thus the outcome of the philosopher's game in such cases is of no value to the scientist. If this is so, then Hofweber's proposal appears to be that philosophers should attempt to save ontology by finding questions in which scientists will admit some interest, and then produce scientifically uninteresting and irrelevant answers to these questions. This is indeed a modest proposal, but it is also conformant with the traditional view that the philosopher's role, with respect to science, is that of a commentator who should risk no conflict.

[3] (Jacquette, 2002) For the initial statement of the problems Jacquette seeks to address, and the distinctions to be employed, see pp. 1-11.

In traditional philosophy (and in philosophy traditionally practiced by contemporary philosophers) there is then a general recognition of the importance of ontological questions in science together with widely held views that ontology (and hence the addressing of ontological questions) is an essentially philosophical task. But philosophers seem to disagree strongly concerning whether there is a role for the philosopher on the scientific side of ontology, and there is little insight offered concerning exactly what this role is or precisely how it is to be played. A number of different models are proposed of the relationship between science and philosophy, and between scientist and philosopher.

Russell endorses what is fundamentally an *evolution model* in which, as science advances and fragments of philosophy break off to form special sciences, ontological work (at least in a particular science) becomes the work of the scientist. Such a model is also explicitly urged by Barry Smith (cf. Smith 2009) who takes psychology to be a paradigm case of such evolution. But in Smith's case — playing directly to the sort of fears expressed by Hofweber — he sees ontology itself breaking off as a special science. A consequence of this model is that the connection between the philosopher and scientist becomes obscure in the process of evolution and speciation, and at the very least there is a movement of certain philosophers and a certain domain out of philosophy and into either its own or another discipline.

Hume embraces a rather vague *diffusion model*, and what is missing in this case is any hint of a mechanism by which the philosopher may affect science. That the philosopher's work will have an effect in science then appears to be more an expression of faith than reason. The same is true for the sort of *grounding model* proposed by Jacquette which argues for a strong *logical* connection between pure philosophical ontology and applied scientific ontology, but which again falls short of characterizing any *practical mechanism* by which the one may affect the other.

Hofweber's approach may most charitably be described as acknowledging the accuracy of Russell's evolution model while invoking an *overlap model* according to which philosophers may keep their hands in ontology to the degree that both scientists and philosophers are interested in a certain core of ontological questions and answers to them. But there are some significant flaws in this view since the kind of interest had by philosophers in these questions (e.g., Hofweber's common example of "Numbers exist") is quite different from the interest that scientists have in them,[4] and again any way in which what the philosopher has to say would be of interest or relevance on the scientific side of an overlap case is left wholly obscure. The questions that Hofweber finds to be of interest and to appear as overlap cases are ones whose answers (whatever these may be) will not contribute in any way to the progress of science.

Finally, Leibniz and Peirce rather aggressively support a *participation model* according to which the work of the philosophical ontologist is directly relevant to that of the scientist. But here again the details of any mechanism of participation remain absent.[5]

---

[4]  Jacquette certainly sees this difference quite clearly, and his distinction between pure philosophical ontology and applied scientific ontology is largely addressed to it.
[5]  We can, however, perhaps take the examples of the lives and work of Leibniz and Peirce as illustrations of such participation.

In all this, what is left undetermined is exactly how the philosopher will affect the science, how what the philosopher does (which presumably is *philosophy* of one sort or another) can contribute to what the scientist does (which presumably is *science*), and by what process this contribution may be accomplished. Are we to hope (as Hume seems to suggest) that scientists will read the abstract work of philosophers and somehow come to apply this to their scientific work in constructing and testing hypotheses and theories? Or is there a more direct way, as Leibniz and Peirce suggest, in which philosophy and philosophers may influence science?

## 1. Ontology, science, and data

The customary view of the philosopher in relation to science is that of being a commentator on science, on the meaning (or "grounding") of scientific claims and theories, and on the methodologies of science. Such a perspective is compatible with the views of metaphysics discussed above (it is indeed expressed succinctly by Russell and is evident in Hofweber and Jacquette), and it essentially eliminates the philosopher from *participating* in science. Philosophy, according to this view, helps us to understand science and how it works (or how it does not, in the cases where it does not or has not), and some portions of philosophy (e.g., logic, inductive logic, epistemology associated with statistics) may contribute in some way to the methodology of science, but the philosopher does not participate directly in the scientific enterprise. Scientists do science; philosophers do not.

Historically, one of the forces driving a wedge between science and metaphysics (or more generally philosophy) was the development of technology in the late Middle Ages and Renaissance that allowed for the design and execution of more careful empirical investigations. This led to an enhanced ability of science to make more accurate and reliable predictions, which in turn led to complex conceptual schemes (scientific theories), supported by experiment, that enhanced this ability even more. In short, the result was the "definite answers" and "knowledge itself" to which Russell and Taylor refer: scientific knowledge.

Some of this knowledge was incompatible with the views previously expressed by the "great thinkers" such as Aristotle, and thus unfortunately this trend resulted in confrontations (as in the case of Galileo) with other strong forces until a kind of armistice was declared in which science and philosophy were agreed to have distinct domains. A consequence of this over time was a separation not simply of the disciplines of science and philosophy, but of scientists from philosophers in the sense that scientific work was done by scientists and philosophical work was done by philosophers. While some scientists might be philosophically sophisticated and some philosophers scientifically astute, a definite division of labor and separation of methodologies arose that resulted in the extinction of the philosopher/scientist or "natural philosopher" of ages past. This divide widened over the centuries as the volume and diversity of scientific knowledge increased dramatically, and science itself fragmented into an increasing number of sub-disciplines. It has left us with the view expressed by a number of philosophers that while ontology may be relevant in some way to science, the philosopher's work in ontology must remain pure and abstract while the scientist's needs and work — pertaining to what Jacquette at one point refers to as "motley existence requirements" (Jacquette 2002, p. 6) — are not the sort of thing that the philosopher has either a right or an interest in pursuing. Can we retain such a view in the face of contemporary science?

When we look at the state of ontology today and its relation to science, a number of questions naturally arise. To begin: What has caused such a dramatic shift from ontological research in philosophy (where this has always been merely a sub-area of metaphysics with some interest to logicians, philosophers of language, and philosophers of science) to the substantially greater time and effort being devoted to ontological research (including funding, large research projects, publications, conferences, societies, and the creation of new journals) that appears to be largely, if not completely, outside of institutional philosophy and firmly within such domains as computer science, information science, and the various empirical sciences? How have we come to a situation in philosophy where the editor of *The Monist* is giving talks on ontology titled (Smith 2009) "Why I Am No Longer a Philosopher"?

Again, as was the case in the late Middle Ages and Renaissance, the answer rests on the advance of technology — this time in the form of digital computers. But while digital computers have been with us for some time (at least since the 1940s), it is only within the last decades of the twentieth century that they acquired the capabilities to support the creation and use of what can only be thought of as large knowledge and inferencing systems. The storage capabilities themselves have exploded; and as the cost of data storage has come down while access speed to the stored data has gone up, it has become trivially possible for scientists to move from collecting kilobytes (thousands of bytes) of data to collecting megabytes, gigabytes, terabytes, petabytes (thousands of trillions of bytes) and beyond — and to design sophisticated software methodologies and systems for extracting knowledge from such masses of data. The availability of such data has resulted in vastly expanded horizons in scientific discovery for astronomers, chemists, physicists, and researchers in the biological and medical sciences.

The change that advances in digital computation and related computer and information science (including algorithms, data structures, database systems, wide area networks, and artificial intelligence) have imposed on the practice of science is emphasized by considering that in 1944, near the end of "The Semantic Conception of Truth", Tarksi attempts to defend his work against the objection that semantics (as he has developed it) is not applicable to the empirical sciences. After some hand-waving, reminiscent of Hume's remarks, about how abstract theories in general, and semantics in particular, may have indirect influences difficult to assess or predict, he concedes that semantics will not have any direct applications in "the natural sciences":

> It is perhaps unnecessary to say that semantics cannot find any direct applications in natural sciences such as physics, biology, etc.; for in none of these sciences are we concerned with linguistic phenomena, and even less with semantic relations between linguistic expressions and objects to which these expressions refer. (Tarksi 1952, pp. 37-38)

Viewed from our contemporary perspective some half-century later, this must be regarded with astonishment since semantic considerations have become central to the representation and analysis of scientific data through the use of large computer systems. Tarski, it would appear, could not have been more wrong; but he also could not have foreseen the effect the combined hardware/software revolution would have on the practice of science. Is there, or should there be, a similar effect on the practice — and teaching – of philosophy?

## 2. Towards a participation model
The collection, organization, and use of data, which lies now at the very heart of empirical science, cannot be accomplished without equally sophisticated systems of classification — which is to say *ontologies* — integrated into the software systems that manage, analyze, and interpret the data. And ontologies, well and efficiently constructed, thereby serve as essential components in the engine of contemporary science. If, as Richard Taylor suggests, metaphysics should be centered on problems in understanding the data of the world around us, then it is clear that metaphysicians should contribute to the understanding of the substantial amounts of data now within the domain of empirical science. But how?

This question may be answered, in the spirit of Leibniz and Peirce, by adopting a *strong participation model* regarding the relation between the philosophical ontologist and science. Such a model is not incompatible with at least some other models of the relation between philosophical ontology and science, and in particular it is quite compatible with the diffusion model of Hume and even the overlap model of Hofweber — though it goes beyond these both in assigning responsibility to the philosopher for addressing certain ontological issues and in the significance and effect that philosophical ontology will have as a result. It is at least somewhat incompatible with the evolution model of Russell and Smith in that it requires philosophers to assume that responsibility rather than abandon it to other disciplines and traditions.

The participation model is also incompatible with the grounding model of Jacquette — not in the sense that it denies the *importance* of the sort of grounding that Jacquette sees as necessary, but in the sense that it denies the *necessity* of establishing such a grounding as a *pre-condition* of doing meaningful and successful ontology in a scientific context. Jacquette feels quite strongly that the role of philosophical ontology is to answer "fundamental questions" of the meaning of "exists", and that arriving at the "right answers" to metaphysical puzzles requires that "philosophical ontology precedes scientific ontology" (Jacquette 2002, pp. 275-276).

In point of fact, science has done very well in the absence of a broadly accepted definition of "exists", and we can continue to expect it to do so in the future. What the philosophical ontologist has to contribute to science is not a single foundational definition, but rather methods, skills, and experience in constructing complex systems of entities, concepts, and languages. The participation model I am advocating here is not one in which the philosopher brings to the scientific table — as a pre-condition of participation — an acceptable philosophical analysis of the concept of being to then be applied in the development of scientific ontologies (as Jacquette suggests), but rather the philosopher brings a set of concepts, skills, and methods that will both inform and be informed by the development of such ontologies.

Philosophers may indeed provide a grounding of fundamental concepts and terms (such as "exists") in a systematic manner as Jacquette believes he has done. Such an exercise may yield useful methods or principles that can be applied to the creation and analysis of ontologies in science. But although such a grounding *system* (whose role is to provide justification and support for a particular philosophical/scientific approach to ontology and ontologies) may be of some pragmatic value in approaching applied ontology, a grounding *model* (whose role is to provide principles and guidance in developing applied ontologies) cannot serve as a unique or necessary guide to the philosopher's participation in science. Science cannot, and will not, wait on the delivery of such a grounding system; and the philosophical ontologist should not withhold participation in science for lack of such a system (or in Jacquette's case, in particular, lack of a definition of "exists"). Science cannot wait for the completion (and presumed acceptance) of such a system prior to creating the "preferred existence domain" that is to be grafted onto that system and that it needs in order to progress.

Moreover, no one can doubt that in fact philosophers will produce (as they have produced) multiple grounding systems of this sort, some incompatible with others. How, then, would the scientist choose which of these to employ as the foundation for the preferred existence domain to be created and grafted thereto? It matters, since there almost certainly will be incompatibilities in grafting to different bases of this sort. Such a concern is expressed by Colomb and Weber when they observe that

> We as information systems researchers are not central players in the effort to understand meaning – we must adopt and adapt theories of meaning from these researchers. Since there are many different and strongly argued positions, if we select one and build on it, we run a serious risk of making the wrong choice. (Colomb and Weber 1998, p. 213)

And this indicates that the role of the philosophical ontologist in science (certainly from the perspective of the participation model) is *not* to serve up a complex metaphysical theory as the starting point for the construction of a scientific ontology, but rather to work together with scientists in applying *effective methods and principles* (based, perhaps, on one or more such theories, even if these are of an incomplete or fragmentary nature) to specific *problems* facing the scientist. In taking a more pragmatic approach of this sort, the participation model rejects any requirement for the completion and acceptance of a grounding theory prior to the philosophical ontologist's participation in the work of applied scientific ontology.[6]

Filling out the details of this participation model requires answering several additional questions concerning the types of problems to be addressed, the warrant or authority that the philosopher has in addressing them, and what the consequences of all this are for the practice and teaching of applied ontology from a philosophical perspective.

## 3. Philosophical problems in applied ontology
What are some examples, in the context of scientific ontology, of questions or problems that can require philosophical attention? And how do these determine the nature and scope of the philosopher's participation? Limitations of space prohibit an exhaustive, or even very detailed, description of the problems in scientific ontology that require the skills and experience of the philosophically trained, but we may at least sketch an overview of the types of such problems and mention briefly one or two specific examples.

There are, in fact, two broad types of contribution that the philosopher may make. The first of these involves the design, analysis, and criticism of specific ontologies; and it falls under the descriptions of such system-building activity as seen in Leibniz and Peirce. It thus involves primarily the application of various methods and criteria to create, to improve, and to evaluate ontologies. Here the philosopher works with scientists either to create a new ontology in a particular domain, or to modify, extend, or repair an existing ontology. For example, the Disease Ontology (see Chisholm et al 2008) was created in 2003 in an attempt to provide a hierarchical representation of human diseases and to describe the relations among the diseases so represented and diseases and medical conditions in other ontologies and terminologies. It has been revised once and is currently undergoing additional active revision in order both to eliminate problems that were discovered in its content and organization , and to make it more compatible with the Unified Medical Language System (UMLS) (see NLM 2010a).

---

[6] There is an associated danger here that insistence on a grounding model and grounding system invites, and that is the temptation for the philosopher to become ideological, dogmatic, and coercive in the appeal to such a system, leading to a kind of philosophical arrogance that steps beyond the bounds of the genuine authority the philosopher has in the domain of applied scientific ontology. For an example of criticisms of cases in which philosophical ideology can be counter-productive in this way, see Merrill 2010a and 2010b.

More recently, some questions concerning the assimilation of the SNOMED CT ontology into the UMLS have been raised; and these involve issues in the representation of one ontology in another, the nature of *concepts* as these appear in concept-based ontologies (such as SNOMED and the UMLS), and relations of synonymy or similarity in meaning that may be used to establish a correspondence between ontologies.  An examination of such cases quickly demonstrates that the issues are not of a purely technical nature (that could be addressed within computer or information science), but rather involve fundamental philosophical questions concerning the relation of one conceptual scheme or ontology to another, how concepts should be characterized, and how two concepts may be related to one another if they appear in disparate complex systems.[7]

The second type of contribution that the philosopher as ontologist may make to science falls within the methodological arena and may arise from our noticing that ontology-specific tasks often assume the existence of principles to be used in completing them. As an example, our task might be to evaluate two ontologies purportedly comprehending the same domain, and with the goal of deciding whether to accept both, accept one but not the other, reject both, or merge the two into a single more acceptable or useful ontology.  But if we are to do this in other than an *ad hoc* manner, we need some principles to guide our analysis and decision.  Failure to find such a set of principles on which to base ontological analyses will result in any decisions or recommendations being a continuing source of dispute, and this in turn will delay or inhibit scientific progress.[8]

---

[7]  NLM (2010b)   Gail Larkin of WebMD raises her question in "Any commentary on synonymy/mapping of SNOMED CT to ICD10PCS concepts?" on June 1, 2010; and Kevin Coonan of the Dana-Farber Cancer Institute posts a related concern as "General questions about mapping SNOMED (and other terminologies) into the UMLS Metathesaurus" on June 15, 2010.  For a thorough treatment of the related notions of concepts and synonymy in the UMLS, see Merrill 2009.

[8]  One illustration of this occurred in January, 2008 in a discussion and dispute on the Open Biomedical Ontologies discussion list (Ashburner et al 2008)  pertaining to principles being proposed for the inclusion of an ontology in the OBO Foundry (basically a library of "approved" ontologies).

The principle in dispute was one according to which the Foundry would permit the inclusion of *at most one* ontology for a particular domain (e.g., biosequences), and that any subsequent competing ontology would need to be "merged" with that one.  The dispute then centered around precisely what "merge" meant and whether, both in principle and in practice, such merging was always possible.  This in turn led to questions concerning the concept of "overlapping" among ontologies, what it meant to say that one ontology was "better" than another (in a given domain), and criticism of the lack of clear definitions or accounts of these concepts.  At one point a participant suggested that the problem would be clarified when everyone realized that "Ontology domains overlap when terms in the ontology have the same meaning".  But it was not generally accepted that reduction of ontology overlap to sameness of meaning (not to mention the use/mention confusion this suggestion involved) lent the desired degree of clarity to the These potential contributions can be seen to fall within a range of different types of ontological problems, each of which is philosophical in nature, and most of which manifest themselves both as problems pertaining to specific ontologies and to methodological considerations in ontology. The problems include problems of content (what should be in the ontology and what should not), problems of organization (how are elements of the ontology related to one another and to scientific data), complexity and its reduction, problems of individuation, problems of commensurability (of one ontology with another), relations of an ontology to its representation language(s) and to the observational language of data, and problems of the adequacy and validation of ontologies (what does "adequacy" even mean, and how and in what sense may an ontology be validated for use in such areas as biomedicine or drug safety?).

A significant set of problems, for example, surrounds questions pertaining to the comparability and commensurability of scientific ontologies, and the possibility of "matching" or "mapping" one ontology to another. Currently proposed criteria for the inclusion of an ontology in the Open Biomedical Ontologies Foundry depend upon the ability to determine that two ontologies overlap or characterize the "same domain", and important questions arise concerning what it means for one ontology to be "better than" another in a given domain, whether and to what degree two ontologies may be "merged", and whether a methodological goal should be to strive for a single "convergent" ontology in a domain or rather to permit or to encourage multiple ontologies that may be incompatible with one another. The problems and questions here are not purely formal or technical, but invite analysis and explication on the basis of such classic metaphysical orientations as realism, conceptualism, and nominalism. And answers to these questions can have direct consequences for what is regarded as "good ontology" in science, what is hence regarded as "good science", and what research proposals my be supported by funding agencies.[9]

## 4. Ontological skills and the philosopher's role

Having seen the problems in scientific ontology that should be of interest to the philosopher and to which philosophy may hope to make significant contributions, we are immediately confronted with the question concerning by what warrant or authority the philosopher may seek to participate in the practice of science in order to propose solutions to such problems. In short, why should we expect scientists pay any attention to philosophers at all?

Part of the answer to this question is that, with respect to contributing to scientific methodology — in the contemporary context of large computer, database, and software knowledge systems — the philosophical ontologist is in a position no different from that of the mathematician, statistician, or

---

discussion. Moreover, as a consequence of the debate it became obvious to at least some of the participants that certain fundamental positions in metaphysics and philosophy of science (e.g., various sorts of "realism") could quite dramatically affect both methodological principles and policies that one was inclined to adopt or to proffer as requirements in the context of practicing science. For a similar illustration, see the exchange in OBO 2010 that was stimulated by Dumontier and Hoehndorf 2010 and Merrill 2010a.

[9]     For some insight into issues pertaining to the characterizations of scientific ontologies, their comparability and commensurability, and how an appeal to philosophical positions may be relevant to such concerns and their consequences, see Euzenat and Shvaiko 2007. Merrill 2008, 2009, 2010a, 2010b, Smith and Ceusters 2010, Dumontier and Hohendorf 2010, Lord and Stevens 2010, and Kutz et al 2010.

computer scientist. In the case of each of these disciplines there is a certain core knowledge (pertaining to useful concepts, theories, techniques, and methods) that is applicable to empirical science and without which empirical science cannot function. Mathematics and statistics (with the related field of probability theory) have been playing such a role for centuries. But only within the latter half of the twentieth century has computer science developed as a distinct discipline, and only within that period has it had an effect on the practice of science — to the degree that it is now inconceivable that science should progress without it. If mathematics is the language of science, then computer science has become its engine.

However, these are formal disciplines and so it seems natural that they should be applied to the problems of science. What about philosophy? Philosophy is one of the humanities, and not a formal discipline at all. How can it hope to contribute to science? But this view of philosophy is both too narrow and a significant misperception. While a number of areas of classical philosophy cannot reasonably be characterized as being of a formal nature, certainly others can. Metaphysics in the tradition of western philosophy, and particularly ontology, is one of these, and is substantially a formal discipline — even if its theories and analyses have traditionally been expressed in natural language. Certainly Aristotle's goal was to formalize — in one way or another — what we know and how we know it, and as part of this, what there is and how it is organized. In this respect philosophy (and particularly ontology) has no less claim to scientific relevance than do mathematics, statistics, and computer science. And in order to have its broadest and deepest effect on natural science, computer science needs philosophical ontology no less than philosophical ontology needs computer science. In the crucible of contemporary empirical science, computer and software systems serve as laboratories for ontological theories and methodologies, lending empirical content and practical effect to those theories and methodologies.

Somewhat oddly, philosophers have never been particularly adept at elucidating any special skills they may have. In discussions of philosophical skills, the first mentioned are always logic, the construction and criticism of arguments, and analytical skills (usually in conjunction with "conceptual analysis") in a sense that is never made abundantly clear. Then reference is made to the ability to read and understand, to communicate, and to solve problems. The litany of these skills, alas, does not appear to be distinctively philosophical. What is it about the training of a philosopher that is distinctive and provides him or her with the sort of skills and the access to methods that are of critical importance in contributing to the use of ontologies in the sciences?

Regardless of any particular interest in philosophy (be it metaphysics, logic, ethics, epistemology, philosophy of science, etc.), in acquiring one's *bona fides* a substantial amount of time is devoted to the study of the history of philosophy and of a broad variety of philosophical *systems*. Why is this so? Physicists, chemists, and biologists study little of the history of their disciplines. How many chemists, for example, know what phlogiston is and what were its properties and the experiments confirming these? How would such knowledge, in the normal course of science, help a chemist? Scientists, as part of their education and training at least, do not study their mistakes. Why devote time to studying concepts and theories that were *wrong*? Science advances, and with it, scientists. Yet philosophers dwell on their mistakes (or at least the mistakes of their predecessors and their contemporaries). And *all* philosophers know (to at least some degree) Plato's theory of forms, what its role was, what problems it addressed, difficulties that it presented, and how Aristotle's approach was a response to these. Again, why devote so much time and effort to the study of analyses and theories that are conceded to be *mistaken*? But the answer is simple: because this is where the skills come from. And this is why scientists — unless they are trained also specifically as philosophers — in large part lack those skills.

Ontology is oriented towards the definition, characterization, and solution of problems. Virtually all philosophers agree on this, and it is clearly expressed by Russell, Taylor, Leibniz, and Peirce. It leads to the development of methods and the definition of a set of problems, some of which have been enumerated above in relation to the use of ontology in science, and others which are well-known to philosophers (such as the problem of universals or the problem of change). In this regard, ontology is similar — in its relation to the sciences — to mathematics. The domain of ontology has to do with what exists (with the very concept of existence) and how existents are related to one another. *An* ontology then functions as a *model of reality* (or a significant portion of such a model); and to a scientist, ontology can be seen as *the modeling of reality* — or at least an ontology can be seen as an essential component of building a model of reality in the same way that mathematical models are models of reality. Moreover, the ontology underlies the mathematical models and provides them with an interpretation and meaning they otherwise lack.[10] Ontological modeling in science (often confusingly referred to as "conceptual modeling" by computer and information scientists) is then more fundamental than mathematical modeling since its result is the basic structure to which mathematical modeling is applied and on which theories are built.

Consequently, one way of accurately describing skills that philosophers have in ontology and metaphysics — skills that are recognized by scientists as relevant to their own work — is to say that these are skills of modeling, model construction, and the analysis, comparison, and criticism of models. Such a description is offered by Guarino and Musen in the inaugural issue of *Applied Ontology*:

> The advent of model-driven architectures in software engineering, of model-based approaches for information integration, and of terminological standards for the annotation of experimental data in the sciences has brought the notion of ontology to the center of attention in a range of disciplines. ...
>
> We find it remarkable that an activity that traces its origins to the work of philosophers who lived more than two millennia ago has become central to the development of modern information technology. We find it exciting to be able to articulate broadly applicable principles for ontological analysis and to see how to apply them in new domains. We believe it is essential to look at the details of how modeling may have been done in particular domains and in particular situations in order to extract those generalizable principles. (Guarino and Musen 2005, p. 2)

But it should not be at all remarkable that the work and skills of philosophers have become central to the development of modern information technology and, further, to the progress of science. The critical skills include those drawn from logic (including modal, intensional, and non-standard logics where applicable), but they also include skills from semantics and the philosophy of language (distinctions of use and mention, theories of reference and meaning, the semantics of names and descriptions), epistemology (confirmation, support, refutation, and counterexamples), philosophy of science (theory structure, theoretical terms, and criteria for the acceptance and rejection of theories), and metaphysics (particularly the problem of universals, the one and the many, identity over time). Those skills come, in part, from the philosopher's study of logic and argumentation; but in greater part they come from study and analysis of the prior work of philosophers in constructing and criticizing competing models of reality across two millennia.

---

[10] This is a commonly recognized problem in the case of statistical models and data mining, and is often referred to as *the problem of the interpretability of data*. For an example of how an ontology is employed to address this problem see Liu et al 2007.

What will be the consequences if philosophers fail to assume this responsibility to participate actively in the development of scientific ontologies? There are several. First, as Hofweber and others fear, the scope and import of ontology in philosophy will shrivel and become of little or no interest outside of philosophy itself. The ontological work traditionally done by the philosophically trained and sophisticated will be performed by others. Meta-ontology will remain within philosophy proper, but the sort of ontological research and contributions championed by Peirce, Leibniz, and even Aristotle will move elsewhere. Beyond this disciplinary consequence, there will be more practical effects because we can expect the preponderance of applied scientific ontology to be done by the poorly trained and informed. Already there is significant evidence of such an effect where the philosophically unsophisticated have based ontological work on a poor or incomplete grasp of fundamentals.[11] Along these lines we should note that within the domain of informatics it has become quite popular to repeat an observation of Andrew Collier that "the alternative to philosophy is not no philosophy, but bad philosophy" (Collier 1994, p. 17), and so there is a growing realization within the scientific informatics community itself of both the value of philosophy and the need for direct participation of the philosophically skilled in ontological work.

## 5. Participating in science

The remaining questions concerning the participation model of philosophical ontologists in science are: What are the details of the mechanism of this participation? And what are the consequences of this for how philosophical ontologists must approach their work and teaching? We may begin by identifying several different types or levels of participation for philosophical ontologists in the practice and teaching of science. First among these is the level of *organizational participation* that requires the philosopher to participate actively in scientific rather than purely philosophical organizations. Some examples here, oriented specifically towards ontology, include the International Association for Ontology and Its Applications (IAOA), the International Conference on Formal Ontology in Information Systems (FOIS), The International Conference on Biomedical Ontologies (ICBO), the National Center for Ontological Research (NCOR), the American Association for Artificial Intelligence (AAAI), the American Medical Informatics Association (AMIA), the European Federation of Medical Informatics (EFMI), ONTOLOG, and the Open Ontology Repository (OOR) — though there are a number of others as well. Interaction with such organizations will not only provide the philosopher with detailed information concerning recognized problems and research projects, but will in addition provide rich sources of further education and opportunities for participation in new or continuing projects of an ontological nature.

This leads to the next level of participation in the form of direct contributions to *research and development* in scientific ontologies, and two avenues of participation are open. First, of course, is the route of refereed publications or presentations at conferences. But the choice of venue is critical here since in order for the philosophical ontologist to have any effect on science his or her work must be seen and understood by scientists. And this means publishing and presenting research results outside of those venues traditionally recognized as appropriate for academic philosophers. In addition to conferences of those organizations mentioned above, examples of journals to be considered for such research include *Applied Ontology*, the *Journal of the American Medical Informatics Association*, the *International Journal of Medical Informatics*, the *International Journal of Metadata, Semantics and Ontologies*, the *Journal of Data Semantics,* and a variety of journals in the areas of artificial intelligence, human/computer studies, computer science, and machine learning.

---

[11] There are now numerous instances of such cases, and criticisms of such attempts. For examples and further literature references, see Ceusters and Smith 2007, Merrill 2008, and Smith 2004.

Publishing in such venues will confront the philosophical ontologist with some unfamiliar challenges. The first of these results from the fact that in general science proceeds more quickly than philosophy in terms of both the pace of research and the dissemination of results. Partly as a consequence of this, and partly as a consequence of the manner in which scientific results historically have been reported, the philosophical ontologist will find it necessary to adapt to a different style of publishing and, typically, to the publication of shorter, somewhat less argumentative, and differently structured presentations.

Additionally, a significant difference in the research and intellectual cultures between philosophy and science is that contemporary scientists now pursue their work almost exclusively in teams, where each member of the team may be responsible for a particular set of tasks or area of expertise. It is relatively unusual for philosophers to collaborate on research (beyond at times co-authoring papers or books), and it is unheard of for philosophers to collaborate to the degree and in the manner that this is demanded by contemporary science. In turn, the different style and venues for publications and presentations required under the strong participation model may present difficulties for traditional philosophy departments and university administrations in evaluating the professional significance of scientifically meaningful work done by the philosophical ontologist. There indeed may be a tendency to regard such work as "not philosophy" or "not of philosophical interest", but at base this may be seen as a tacit acceptance of the evolution model, the grounding model, or the overlap model and a rejection of the kind of substantive systematic ontology traditionally practiced by philosophers.

The issue of collaboration and the resulting multiply-authored publications and research studies are only one example of difficulties in applying traditional criteria for promotion and tenure, for example, to philosophers seeking to pursue the strong participation model.[12] This is the basis for one argument in support of Barry Smith's view that it is time for ontology to — in a disciplinary sense — break from philosophy and found its own independent departments: philosophers, and particularly young philosophers, will not be able to succeed on a professional academic level if they devote any substantial amount of their effort to participation in the area of scientific ontologies. However, I think that there is too much to be lost — both for philosophy and for science — in following the evolution model in this case. But a direct consequence of this is that as we must clearly acknowledge the importance of traditional philosophy in the teaching and practice of scientific ontology, we must as well acknowledge the need for university departments and administrations to recognize the contributions of philosophical ontologists outside of areas and venues traditionally recognized as appropriate for academic philosophers.

---

[12] This is not, it must be conceded, a problem only for philosophy. The problem of multiply-authored publications and the determination of a particular team member's contributions has become so extreme in some cases that in recent years professional scientific organizations have adopted stronger and more explicit criteria for inclusion in an authorship list, and have imposed more severe constraints on the number of authors that may be listed for a single paper. However, while scientific departments are accustomed to dealing with these issues, departments of philosophy and schools of humanities generally are not. Moreover, in the humanities – and certainly in philosophy – requirements for tenure and promotion are often phrased explicitly in terms of "singly-authored papers" where in the natural sciences such papers are relatively uncommon (and in fact the ability to work with colleagues in research and publishing on a regular basis is explicitly valued, if not required).

This last point brings us to the third level of participation for philosophical ontologists in applied scientific ontology: *pedagogical participation*. If the participation model is to be adopted and pursued, then this requires also a change in how ontology is taught and the audience to whom it is addressed since we must acknowledge a responsibility for training both new generations of philosophers and new generations of scientists in philosophical ontology, scientific ontology, and the relations between these. Yet if we look at contemporary curricula in philosophy we find few courses addressed to this need.[13]

The requirements for curriculum change in this regard are several, and we must begin with the realization that there are somewhat disparate groups of students to be served. It is natural to think of graduate students in philosophy, and particularly those specializing in such areas as logic, metaphysics, philosophy of science, and philosophy of language. These students can be presumed already to have a substantial background in philosophy and to be prepared for more specific and directed work in applied ontology. But another rich source of students is undergraduates — in either philosophy or the sciences — who may be interested in graduate study or careers in applied ontology or related areas. Finally, there is an important audience to be found among graduate students or post-doctoral students in the sciences (particularly in computer science, information science, and the biological and medical sciences), and this audience is especially open to acquiring the philosophical sophistication and skills that will aid them in dealing with the creation and use of scientific ontologies.

This realization of the need for educating such distinct groups of students (though they share some common interests and goals) then calls for the creation of specific applied ontology courses that may be offered on a regular basis and that provide or extend the necessary training in philosophical ontology into its applications in scientific ontology. Traditional courses in metaphysics, logic, philosophy of language, and philosophy of science cannot be extended to cover this need; and such traditional courses will still be required in order to provide philosophical ontologists their unique perspective and set of ontological skills, and to serve in a background and foundational manner for courses in applied ontology. But it is likewise not sufficient to retreat to a diffusion model, hoping or assuming that those interested in applied ontology will get their training in traditional courses and somehow figure out how to make use of what they learn there when they turn to participating in applied ontology.  A model for this sort of curriculum expansion in philosophy may be found in the inclusion of programs in medical ethics, business ethics, and engineering ethics in recent decades.

---

[13] Some courses appear at the State University of New York at Buffalo and North Carolina State University. Although the Indiana University School of Informatics and Computing was founded by two philosophers — J. Michael Dunn and Myles Brand — it contains no courses with distinctively philosophical content and does not list philosophy as one of its *cognate area*s (integrated programs of courses outside of the school). Courses in applied ontology — some with philosophical content — appear under the auspices of the Laboratory for Applied Ontology of the Institute of Cognitive Science and Technology in Trento, Italy; but these are graduate or post-doctoral in nature and are not taught by philosophers (though the instructors tend to be philosophically sophisticated).

The concern of Russell and Hofweber is that

> ... the questions that metaphysics tries to answer have long been answered in other parts of inquiry, ones that have much greater authority.  And if they haven't been answered yet then one should not look to philosophy for an answer.  What metaphysics tries to do has been or will be done by the sciences.  There is nothing left to do for philosophy, or so the worry.  (Hofweber 2009, p. 160)

But we can concede the cogency of part of this concern without conceding the ultimate consequence it paints for philosophy and ontology — provided that we understand not only the conceptual, but the practical relationship between the philosophical ontologist and science.  It has always been true (at least since the time of Aristotle) that a significant part of what metaphysics tries to do has been done by the sciences.  But it does not follow from this that it has not also been done by philosophy, nor that the ability of science in this regard is independent of philosophy (and this is true in both a conceptual and a practical sense).  And indeed the work of philosophical ontologists — in direct collaboration with scientists — is now, more than ever, critical to the progress of science.

The strong participation model advocated here is distinguished from other views concerning the relation of philosophical ontology to science in several important ways.  Chief among these, perhaps, is that it retains within philosophy a substantive or systematic kind of ontology — one represented by Aristotle, Leibniz, and Peirce, among many others — rather than abandoning the practice and foundations of such a discipline to other domains, and rather than retreating to the practice only of meta-ontology coupled to problems of "what exists" whose solutions are of no interest outside a narrow community of philosophical ontologists.  Instead, it views applied ontology (including the development of ontological methodology and the application of that methodology) as both a central component of metaphysics and a central component of contemporary science; and it assigns to the philosophical ontologist a significant responsibility in ensuring that the scientific applications of ontology are both adequate and correct.  The strong participation model answers questions about the practical value of metaphysics and ontology, and it has consequences for the practice and teaching of ontology as these are approached by philosophers.  The strong participation model also saves ontology *for* philosophy in a meaningful way that the other models cannot, although its more significant goal and effect is to save philosophical ontology *in* philosophy for the benefit of both philosophy and the sciences.

## Acknowledgements

This paper began  in part as a reaction to a North Carolina State philosophy colloquium in which Thomas Hofweber presented a draft of Hofweber  2009.  But it was also written in the context of Merrill 2008, Merrill 2009, and Merrill 2010a in which I was working out some details and examples of my views concerning the roles and values of philosophical ontology in science.  My overall goal is to encourage philosophers to adopt the participation model advocated here and to become directly involved in the work of scientific ontologies.  The paper was rewritten twice into its current form, and it has benefitted greatly and directly from comments and suggestions of Michael Pendlebury and Wayne Martin who, however, should not be held even remotely responsible for either its content or its attitude.